# Thinking Through Signs: PEEL as a Semiotic Scaffolding for Epistemically Accountable AI-Enabled Research

CLARISSE DE SOUZA, PUC-Rio, PUC-Behring Institute of Artificial Intelligence, Brazil
GABRIEL BARBOSA, PUC-Rio, Brazil
SIMONE DINIZ JUNQUEIRA BARBOSA, PUC-Rio, PUC-Behring Institute of Artificial Intelligence, Brazil
BÁRBARA BETTS, PUC-Rio, PUC-Behring Institute of Artificial Intelligence, Brazil
RENATO CERQUEIRA, PUC-Rio, PUC-Behring Institute of Artificial Intelligence, Brazil
JULIANA JANSEN FERREIRA, PUC-Rio, PUC-Behring Institute of Artificial Intelligence, Brazil

Large language models are reshaping research practice while quietly eroding researchers' epistemic accountability. This commentary introduces PEEL (Protocols for Epistemically Engaged Literacy in AI), a working scaffolding that combines deterministic distant reading via Voyant Tools with LLM interpretation via Claude, grounded in Peircean semiotics and abductive reasoning. Applied to AI-generated condensations of three source texts, PEEL reveals systematic distortions in quantity, term frequency, and epistemic voice that are invisible without non-AI measurement — and yields three design implications: deterministic instruments must accompany AI tools; fluency is not fidelity; epistemic authority must be designed in, not assumed.

CCS Concepts: • **Information systems** → **Document representation**; • **Computing methodologies** → **Philosophical/theoretical foundations of artificial intelligence**.

Additional Key Words and Phrases: AI-enabled research, Qualitative Research, Epistemic Authority, Signs and Representations, Abductive Methods

## 1 Introduction

Conversational AI systems using large language models (LLMs) have transformed how research is conducted — and nowhere more consequentially than in text-based, interpretive inquiry, where reading, coding, and argument are the primary instruments of knowledge discovery. Tasks that once required weeks can now be accomplished in hours, and corpora too large for any individual

Authors' Contact Information: Clarisse de Souza, clarisse@inf.puc-rio.br, PUC-Rio, PUC-Behring Institute of Artificial Intelligence, Rio de Janeiro, Brazil; Gabriel Barbosa, gbarbosa@inf.puc-rio.br, PUC-Rio, Rio de Janeiro, Brazil; Simone Diniz Junqueira Barbosa, simone@inf.puc-rio, PUC-Rio, PUC-Behring Institute of Artificial Intelligence, Rio de Janeiro, Brazil; Bárbara Betts, barbarabetts@esp.puc-rio.br, PUC-Rio, PUC-Behring Institute of Artificial Intelligence, Rio de Janeiro, Brazil; Renato Cerqueira, rcerq@puc-rio.br, PUC-Rio, PUC-Behring Institute of Artificial Intelligence, Rio de Janeiro, Brazil; Juliana Jansen Ferreira, jjansen@esp.puc-rio.br, PUC-Rio, PUC-Behring Institute of Artificial Intelligence, Rio de Janeiro, Brazil.

researcher are now, at least superficially, tractable. The research community has understandably focused on what AI-enabled research can now do, with enthusiasts envisioning the possibility of automating scientists and science [8, 11].

Meanwhile, a quieter, more consequential question has been growing: what does the introduction of LLMs do to the epistemic ground on which research stands? Gopal and co-authors [8] manifest the importance of benefiting from new opportunities without losing control of the mechanisms that advance science: insight and oversight. This commentary responds to the TMIS call by examining both, focusing on risks to oversight and deferral of insight in research that aims to discover knowledge encoded in natural language texts. The concern reaches across research paradigms, since all scholarly inquiry depends on reading what has been established, understanding what it means, and making defensible judgments about what the literature warrants.

The commentary proceeds in five sections. Section 2 discusses our specific framing of Gopal and co-authors' challenges. Section 3 articulates how qualitative research meets semiotics through abductive reasoning — a grounding central to everything that follows. Section 4 introduces PEEL (Protocols for Epistemically Engaged Literacy in AI), a working environment combining deterministic text analysis via Voyant Tools with stochastic LLM interpretation via Claude. Section 5 presents evidence from applying PEEL to texts drawn from our own references, creating intentional reflexivity in the commentary. Section 6 discusses design implications, and Section 7 closes with final remarks.

All of us are experienced qualitative researchers working across software design, human-computer interaction, artificial intelligence, and data visualization. PEEL is not a hypothetical proposal but a working environment we have built and used. The ideas and questions we offer here are, in Schön's terms, the result of both reflection on action and reflection in practice [21].

## 2 The Epistemic Stakes of AI-Enabled Research

Long before the age of AI, Hirschheim, Klein and Lyytinen called attention to the dominance of quantitative approaches in IS development [9]. CAQDAS narrowed the technological gap between qualitative and quantitative researchers, but, as Woods and co-authors noted [23], reflexivity is not granted by technology — it requires awareness, will, and discipline.

LLMs have changed the game, but not the winning targets. Blohm and co-authors [4] recognize the opportunity for contemporary AI to support different kinds of IS research designs, methods, and aims. Yet LLMs surface a magnified version of the challenges that emerged with CAQDAS. Interpretive validity rests on the inquirer's intimacy with what is being investigated, transparent positionality, and consistency in the interpretive steps leading to conclusions — all of which the opacity of AI threatens. Bender and Koller [3] detail the limits of *intelligent* natural language processing — limits that connect directly to a philosophical problem: do researchers who delegate interpretive steps to AIs retain full epistemic authority over their conclusions?

Ferrario and co-authors [7] draw on social epistemology to argue that we must examine a new epistemic subject — a *hybrid* of AI and human — and investigate the conditions of authority in this new context. Alvarado's [1] characterization of AI as an epistemic technology leaves the door open to AI-enabled research, whose complexity Boisseau [5] explores in depth. The major risk, Boisseau warns, is agreeing to a technologically-driven reconstruction of *expertise* — one that undermines scholarly accountability and deteriorates the conditions for critical judgment.

Messeri and Crockett [14] called attention to *the illusions of understanding* that AI brings to research, warning against the equivalent of *monocultures*: when only one kind of plant is grown, the soil becomes exhausted. Only pluralism can keep us productive, and the incentive for creative approaches must be preserved, lest extensive AI use retard solutions rather than accelerate them.

## 3 Abduction and the Logic of Interpretive Inquiry

A large part of qualitative inquiry in IS is informed by Grounded Theory (GT) [6]. A productive contention among GT adopters asks whether GT is exclusively inductive or proceeds in iterative cycles where theoretical commitments orient, and are refined by, data collection. Reichertz [19] and Tavory and Timmermans [22] converge: theory-informed GT is a case of abductive reasoning. Abduction, introduced by Peirce [17] at the turn of the twentieth century, produces *hypotheses* — unlike deduction, which produces *confirmation*, and induction, which produces *facts* — and is the driving force of theory building and correction.

Bringing abduction into the picture matters for two reasons. First, abduction is the logical basis of interpretation — a connection originally drawn by Peirce[18] and later popularized in linguistics by Hobbs, with his account of interpretation as abduction [10]. Second, researchers like Magnani and others have shown that abduction is connected to creativity in science [12]. His concept of *manipulative abduction* is particularly relevant for us: it describes reasoning that happens *through* doing, rather than *about* doing — as when a geometer thinks by constructing and manipulating a triangle rather than merely reflecting on the properties of one. It refers to extra-theoretical behavior that creates communicable accounts of new experience, integrating them into existing systems of experimental and theoretical practice. Since, as Messeri and Crockett warned, creativity may be threatened in contemporary AI-boosted research [14], the time is ripe to explore how the merits of abductive reasoning can be translated into technologies.

## 4 PEEL: A Preliminary Proof-of-Concept Answer

In response to the challenges raised in the preceding sections, we have designed and implemented a semiotically rich technological environment for AI-enabled knowledge discovery from natural language texts. We selected three texts from our bibliography – Alvarado (2023) [1], Ferrario (2024) [7], and Boisseau (2026) [5] – and compared how three different AIs summarized each one, reducing it to 25% of the original size, which is, in itself, a form of knowledge extraction. AI-1 and AI-2 produced their summarizations unconstrained by further instructions. They were plainly prompted to produce a summary of the source text. AI-3, Claude, unlike the others, produced the summarization in a highly controlled way. It used instructions encoded in an elaborate *skill*.

We start PEEL's description with Claude's verdict when asked to compare the three summarizations:

"The three summarization modes differ substantially in the epistemic resources they offer. The free summarizations (AI-1 and AI-2) produce fluent, broadly accurate accounts, but neither indicates which sentences are verbatim, which arguments have been compressed, or which examples were dropped. A researcher relying on either could not assess whether a specific claim was genuinely in the original or an artifact of the summarizer's reconstruction — and would therefore be unable to cite responsibly or detect misattribution. AI-2's prose format, while more readable, is marginally less reliable than AI-1's structured one: it tends to dissolve argumentative fine-grained into smooth narrative, creating a false impression of comprehensive coverage. The skill-guided AI-3 condensations are a qualitatively different epistemic object. The injection-type annotations, verbatim percentages, and cluster-coverage tables function as a rudimentary audit trail, making the summarization's limitations visible and interrogable rather than invisible. This supports calibrated trust rather than unconditional trust. The residual risk is that the appearance of rigor may paradoxically induce over-confidence in passages that are heavily compressed; researchers should treat

```
AI-3 outputs as structured finding aids and verify any claim they intend to cite
against the original."
```

The verdict says more about the effects of our skill on Claude than about the other AIs. For this reason, we keep them anonymized. When skilled, they can produce different results. We are not talking about AIs in general, but about harnessing AIs. In section 6 we provide evidence from Voyant's analyses showing that Claude's verdict is confirmed. We also give readers access to the complete skill that guided Claude's behavior.

### 4.1 PEEL's design rationale

PEEL combines two different kinds of support technologies. One is an AI agent; the other is a non-AI text-analysis tool. Standing between computer tools that explicitly operate *adversarially*, with one's strength being almost always the weakness of the other, the researcher is in a privileged and intentionally asymmetric position to decide when, where, how, and why to use, combine and contrast the results of both tools, as the inquiry unfolds.

As already mentioned, PEEL is implemented with Anthropic's Claude[1] and Voyant Tools,[2] a mature, widely tested platform for linguistic analysis and distant reading [16, 20]. The former carries the virtues and costs of stochastic computing [2, 3]. The latter offers the stability and limitations of deterministic natural language processing [13]. While Claude offers a full-blown natural language chatting and prompting interface, Voyant provides rich interactive tabular and visual representations of text analysis, in addition to a literate programming API called Spyral. A *Spyral Notebook* is the key representation of interpretive research activity in PEEL. It records and operationalizes Claude's and Voyant's text analysis outcomes. The researcher works on this notebook, modifying and expanding it, with the possibility of sending it back to Claude for meta-analysis.

### 4.2 PEEL in Practice

We describe PEEL with a single text. Larger numbers of texts can be analyzed and compared, although further investigation is required regarding the scalability of this model.

The process has three phases. Phase 1 begins by uploading the *text of interest* (ToI) to Voyant, which automatically affords the use of all Voyant Tools (see Figure 1). Carefully designed articulation with Claude begins by instructing the AI on how to work with Voyant's tokenization and quantitative analysis. Claude follows instructions to take ToI tokens, along with their Raw Frequency, Relative Frequency, Relative Peakedness, Relative Skewness, and Distribution, and produce: a list of the most significant terms; a list of the least significant terms; and a semantic clusterization of the most significant ones. In order to ground semantic analysis on lexical semantics rather than Claude's inscrutable knowledge base, PEEL uses Wordnet [15]. Wordnet also supports *stemming*, so that morphological word variations can be counted as units in subsequent phases. One of the practical effects of this strategy is word-cloud visualizations with stems, rather than words (see Figure 2). Stemmed-forms semantic clusterizations also allow Claude to generate automatically candidate lists of Voyant categories to be used in visualizations (see Figure 3) and search (see Figure 4).

Phase 2 is where Claude produces *text condensations*, working not only with the source text and the results of Phase 1, but also with explicit discourse and rhetorical strategies included in a specific *condensation skill*, such as:

```
-  Follow the argumentative structure of the source, not its sentence order.
```

[1] https://claude.ai/
[2] https://voyant-tools.org/

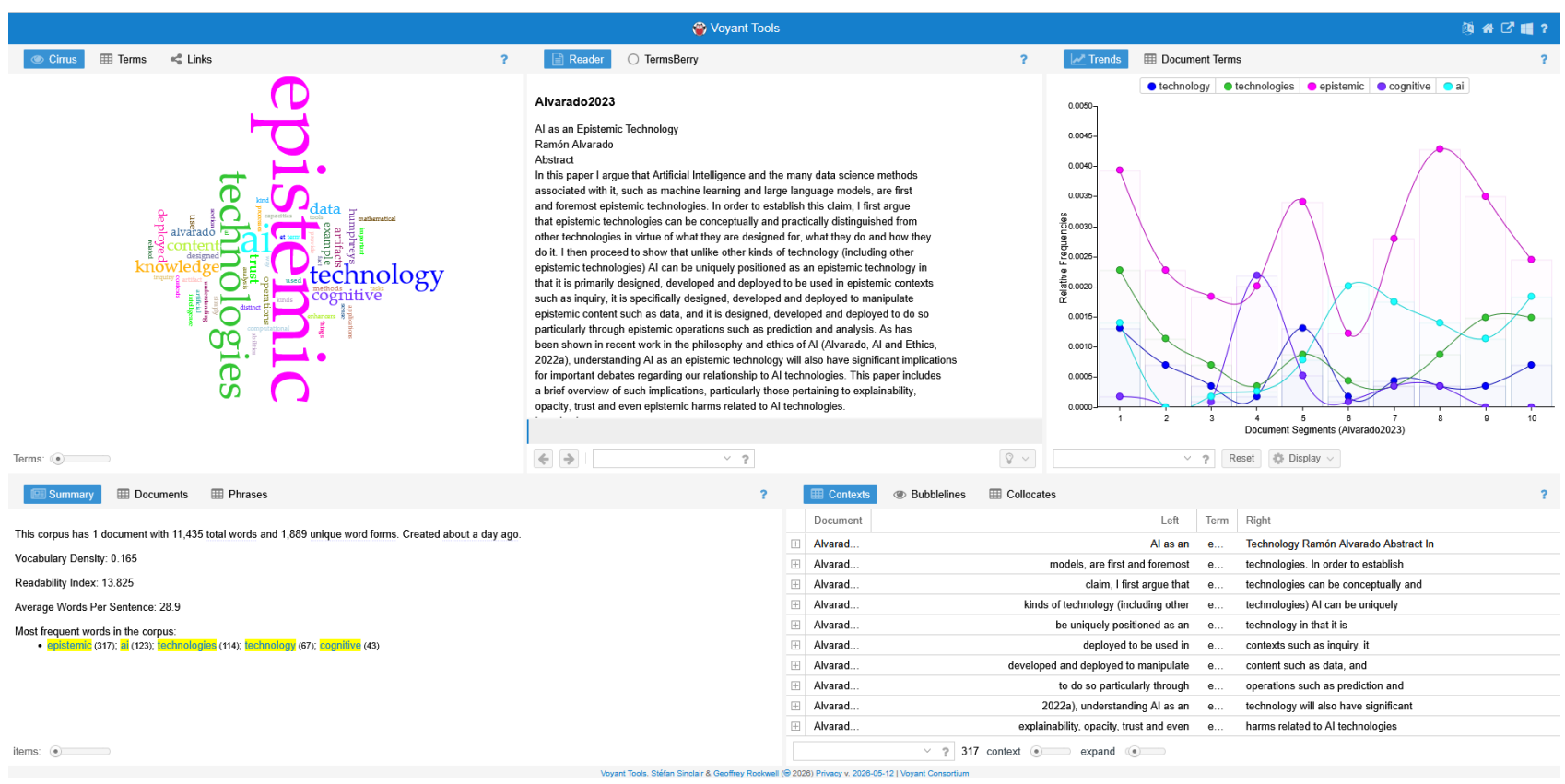

Fig. 1. Voyant Tools Panel - PEEL EXAMPLE (ALVARADO, 2023)

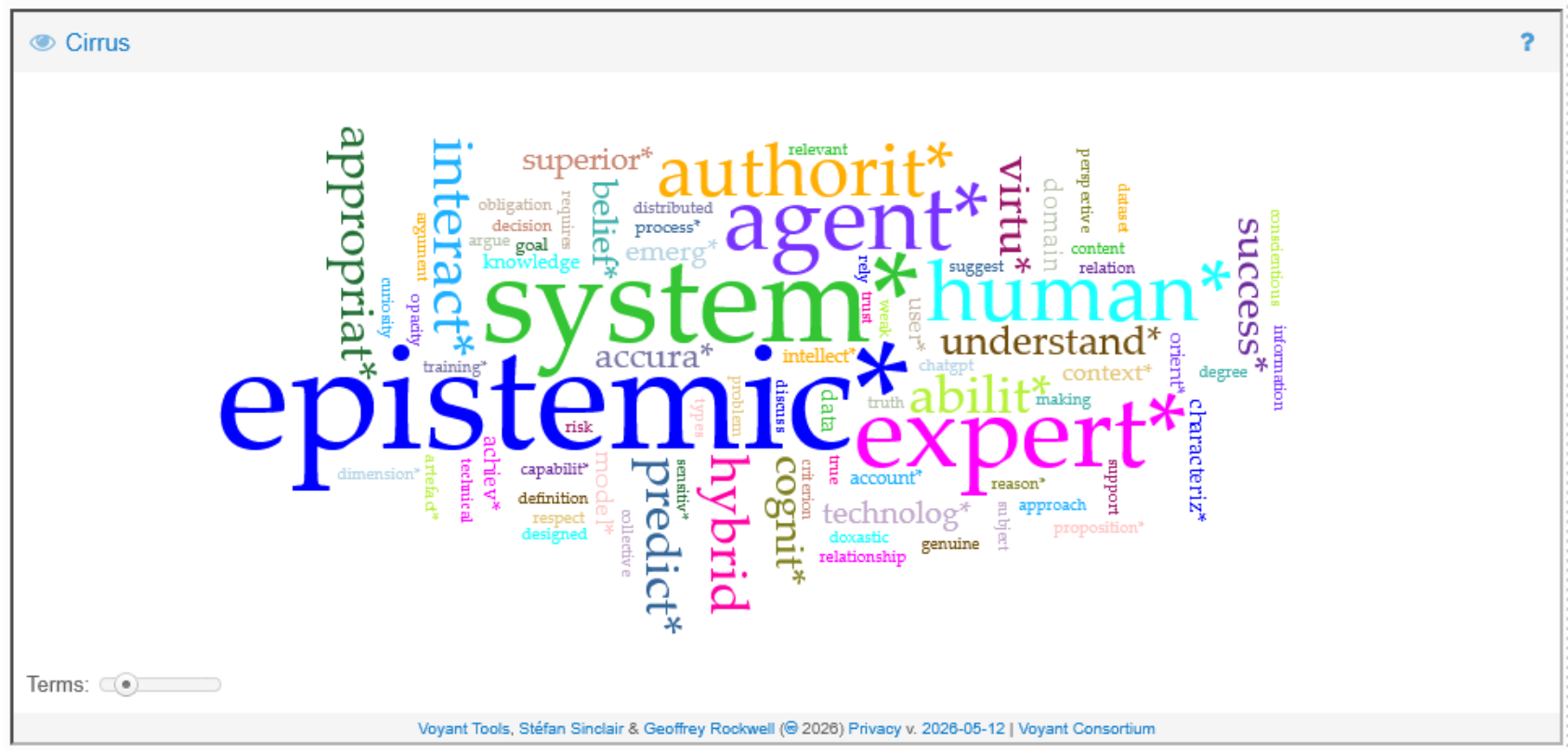

Fig. 2. Voyant Cirrus - PEEL EXAMPLE (FERRARIO ET AL., 2024)

```
Select and compress content by asking: *what does the reader need to
understand to follow the next step of the argument?*
...
- When compressing, prefer to keep the source's own wording for the
  conceptually dense passages (definitions, distinctions, conclusions).
  Rewrite only where the source is redundant or where multiple sentences
  make the same point.
```

As the directives above show, the *skill* is explicitly trying to guide the summarization process to privilege the selection of source material, not only regarding grammar, but also content selection. Further parts of the skill produce a classification of condensed terms. A span is a Framing injection if it contributes no content beyond the immediately surrounding text; a Transition if its subject is the text or argument itself rather than the ideas it carries; a Reformulation if its propositional content can be traced to a single identifiable source sentence; and a Compression if it merges or omits material from multiple source sentences. Since the latter is epistemically risky, the skill explicitly asks the researcher to decide on which risks to take Figure 5. The result of Phase 2 is the annotated *condensation* of text.

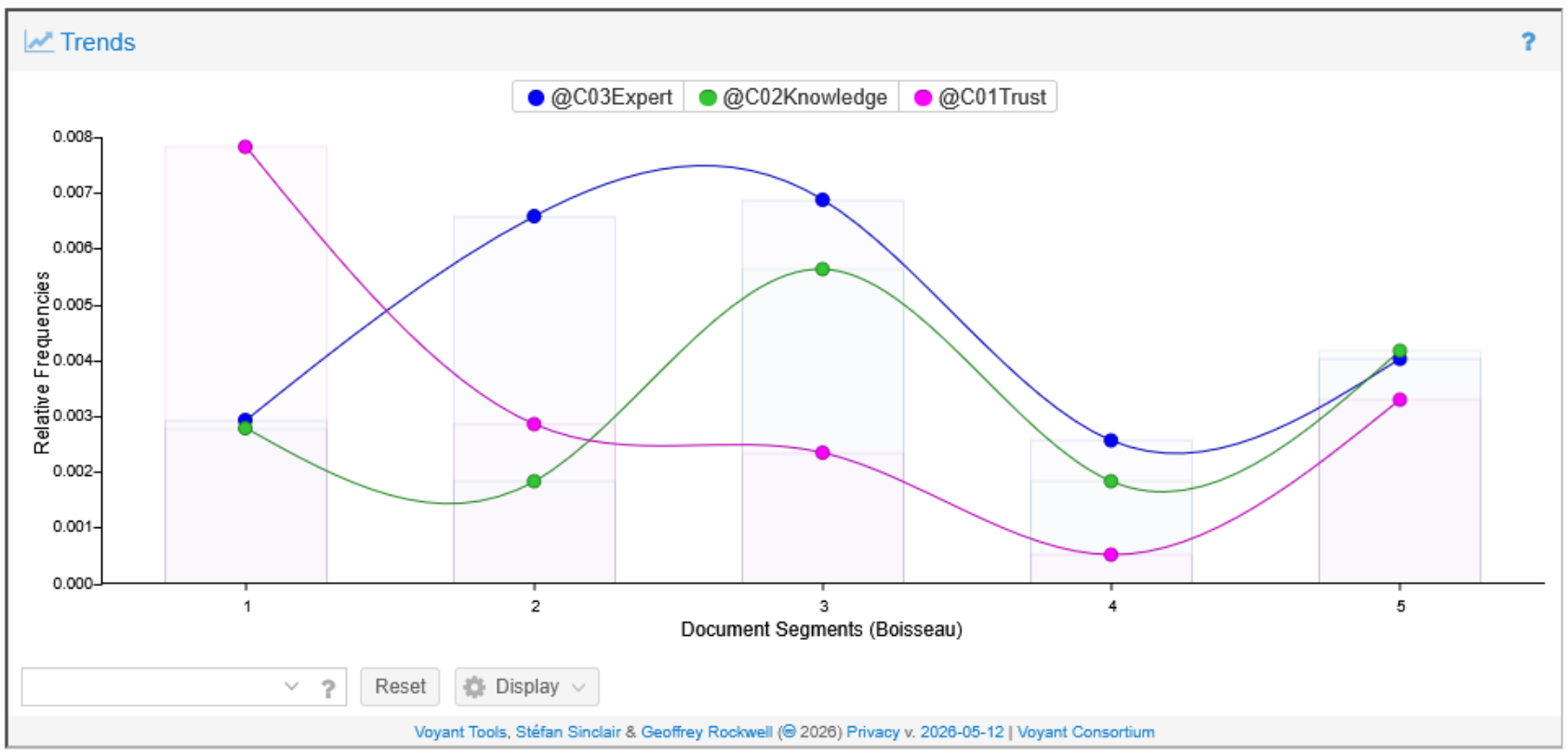

Fig. 3. Voyant Trends - PEEL EXAMPLE (BOISSEAU,2026)

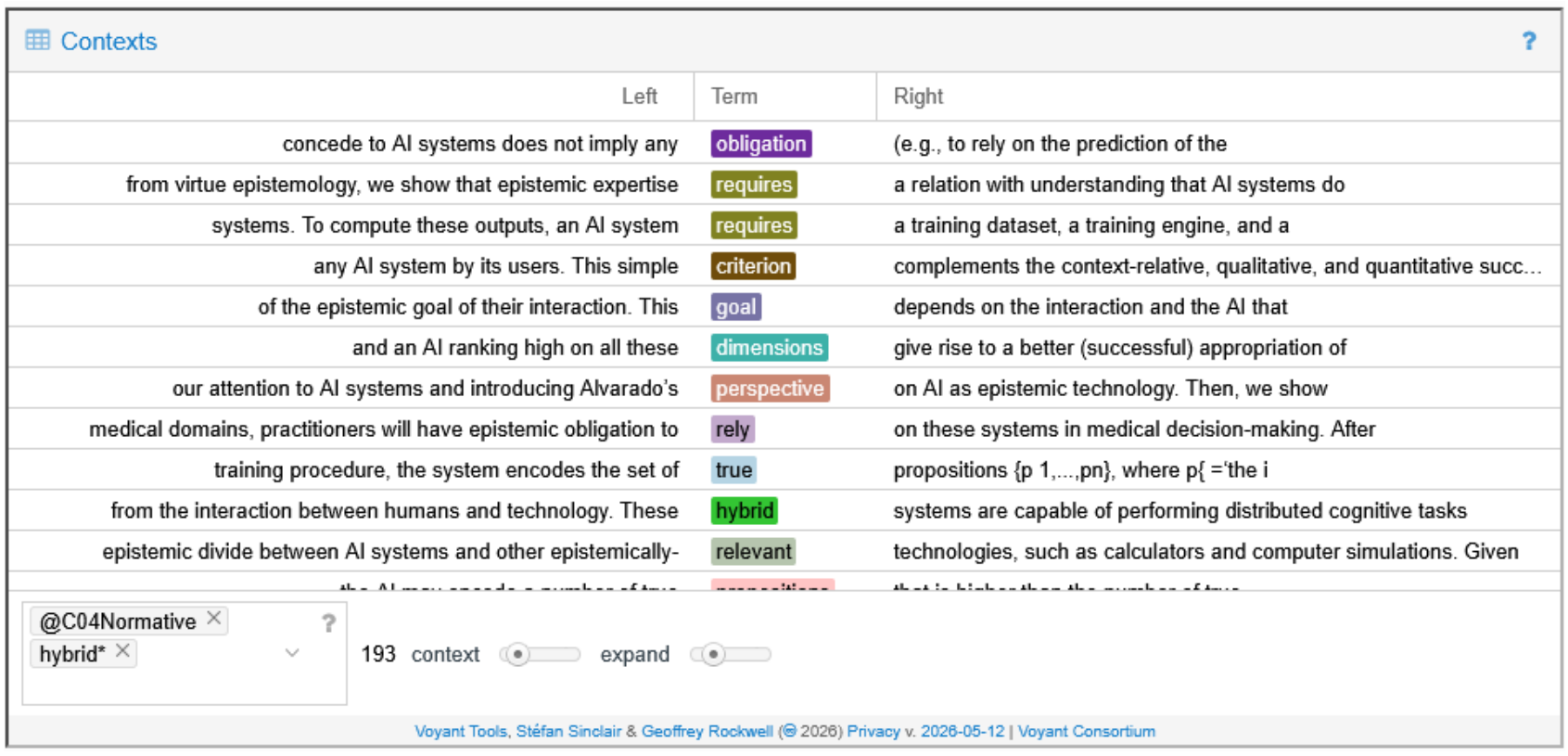

Fig. 4. Voyant Search - PEEL EXAMPLE (FERRARIO ET AL., 2024)

Phase 3 simply takes a model Spyral Notebook and fills in the code cells with JavaScript snippets that translate semantic clusters into Voyant categories, creates the stop-word lists, and generates configurations parameters for ten different Voyant tools (see PEEL Spyral Notebooks examples)[3]. Since Phase 3 is the least dependent on AI technology, we are working on a new version of PEEL where this phase will be carried out using Python scripts, only.

## 5 Evidence from Three Epistemic Encounters

The three papers selected for PEEL's proof of concept are not arbitrary choices. They support a large part of this commentary's argument. What follows reports on Voyant's analyses of three AI-generated condensations of each paper, alongside the source. A disclosure is in order: as part of this study's design, none of the authors had read these papers in full when the Voyant analyses were conducted. The findings motivated the reading, not the other way around. This is manipulative abduction enacted in the researcher's own practice: the tools generated the inquiry; the inquiry generated the understanding. The complete skill used in Phase 2 is available at the bottom of our Spyral Notebooks.

[3]Spyral Notebooks may take time to load and compute, not only because of their content but also because of server traffic.

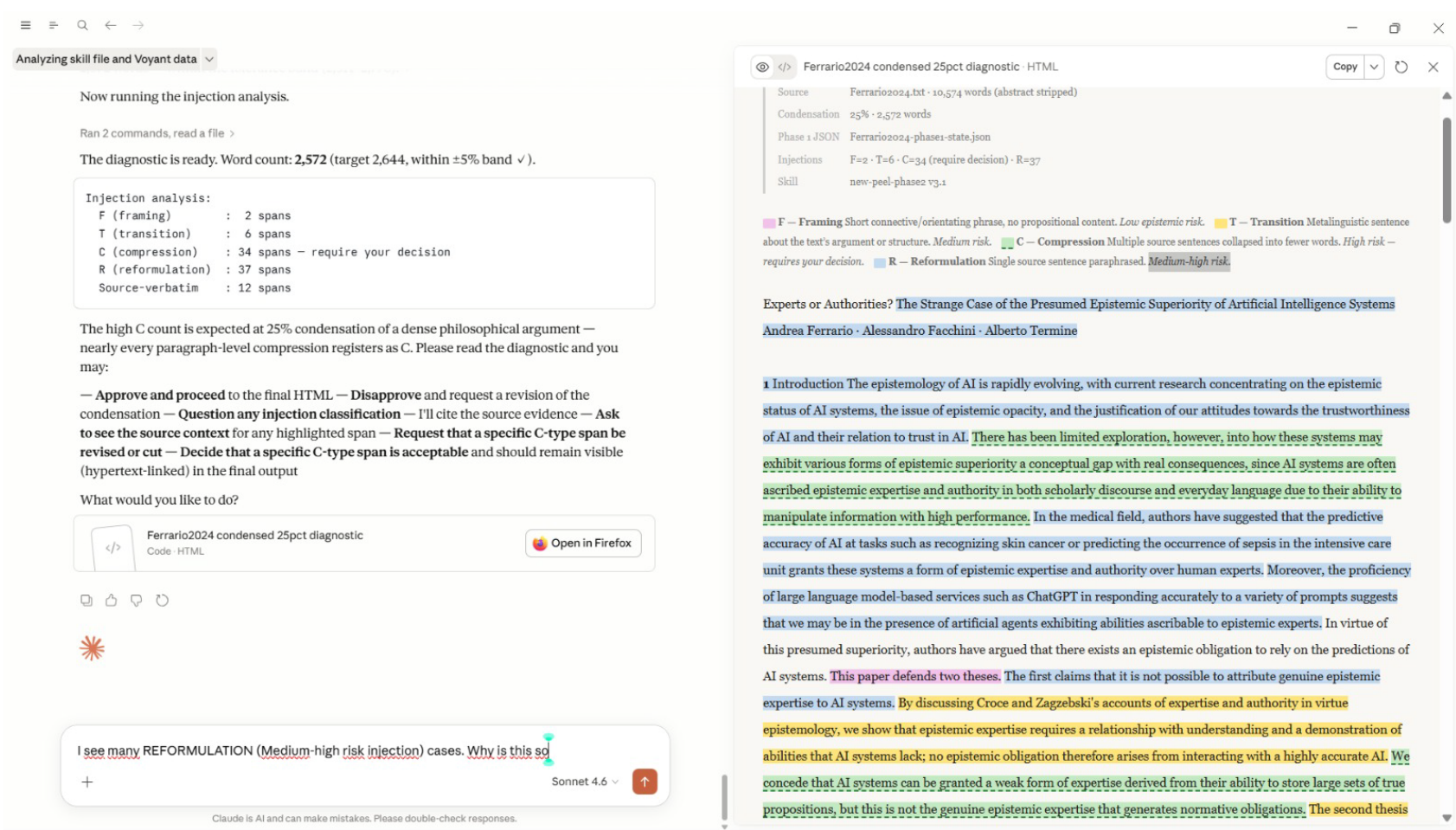

Fig. 5. The researcher discusses the condensation of text with Claude, before approval

## 5.1 Quantity: the condensation task

The instruction to each AI was simple: produce a summary of the text, which is 25% of the original's size, with 5% tolerance. Voyant's Documents tool applied uniform tokenization to all documents in all three sets. The results are unambiguous (check the notebooks with the data). Across all three papers, AI-1 and AI-2 produced condensations of approximately 12% of the original — less than half of what was requested. AI-2's condensation of Boisseau reached only 7.7%. Claude's outputs ranged from 24.9% to 29.7%, remaining in the right order of magnitude and erring upward rather than downward.

The upward deviation in Claude's outputs has two inspectable explanations: metadata and annotation content produced by the PEEL skill inflates; and the researcher's choices at the skill's epistemic checkpoint — where approval must be explicit (see Figure 5) — may have favored completeness over brevity. Both factors are recorded in the pipeline. The deviations of AI-1 and AI-2 are neither explained nor visible. A researcher relying on either would have received a fluent, well-formed text with no indication that it represented half the source material requested.

This finding alone generates four competing hypotheses: computational resource constraints on unpaid subscriptions causing silent truncation; internal compression defaults overriding explicit instructions; silent instruction failure without disclosure; and, for Claude only, researcher epistemic choices recorded at the checkpoint inflating the compression rate. The data cannot disambiguate between them. PEEL makes the anomaly visible and the hypotheses formulable. Without Voyant, there would be nothing to explain.

## 5.2 Meaning: term frequency distortion

Voyant's Bubblelines tool — with "Separate Lines for Terms" enabled, a non-default setting whose activation is itself a researcher choice — reveals a second category of distortion. In Boisseau (2026) [5], whose argument turns on the relationship between trust and epistemic responsibility, the source ratio of `responsib*` to `trust*` is 23:132, approximately 17%. Claude's condensation tracks this closely at 16%. AI-1 produces a ratio of 60%; AI-2, 31%. A researcher relying on AI-1's summary would conclude that responsibility is the organizing concept of Boisseau's paper. It is not. Trust is. The distortion is invisible without measurement.

A further finding cuts across all three AIs including Claude: all three invert the relative frequency of `reliab*` compared to the source. The consistency of this inversion across otherwise divergent systems suggests a systematic pattern in how LLMs process this semantic field — possibly conflating reliability with trustworthiness in ways that distort Boisseau's careful distinctions. This is flagged as a hypothesis for further investigation, not a conclusion. One possibility is that the clusters-to-categories mapping approved by the researcher is not good.

### 5.3 Voice and profile: what the text sounds like

Voyant's Summary tool reveals a third pattern. AI-1 and AI-2 consistently cluster together in their quantitative textual profiles, while Claude and the source cluster together. This is by construction: PEEL is explicitly designed to adhere to the linguistic characteristics of the source text. The free summarizers converge on a generic summary genre; Claude converges on the source.

The most humanly legible expression of this difference is the person of narration. AI-1 and AI-2 invariably write *about* the author, referring to them in the third person: "the author argues," "Boisseau claims." Claude writes *as* the author, in the first person, as the source does. This is not a stylistic detail. In academic argument, first and third person carry different epistemic stances. A summary that says "the author argues X" has already introduced an interpretive distance; it has repositioned the summarizer as an external commentator rather than a faithful conduit. PEEL preserves the author's epistemic voice. The free summarizers replace it with a reporter's.

Finally, the type-to-token ratio — Voyant's measure of vocabulary diversity — is systematically higher in AI-1 and AI-2 than in Claude. PEEL's tighter lexical control is a feature, not a limitation. But it also raises a question this paper plants without answering: is greater vocabulary diversity in a summary a door open to the quiet injection of concepts not present in the source? Careful comparative analysis will tell.

## 6 Implications: Designing for Epistemic Authority

The findings above are not arguments against AI-enabled research technology. They are arguments for designing it differently. Each finding maps onto one layer of the theoretical chain this commentary has assembled. This directly addresses Alvarado's views on AI as epistemic technologies [1]: we must design them with more than epistemic functionality.

The silent non-compliance of AI-1 and AI-2 with a simple quantitative instruction is Boisseau's opacity problem in its most elementary form. The researcher issued a clear, verifiable request. Two systems failed to honor it without disclosure. Without a deterministic, inspectable counterpart — here, Voyant's Documents tool — the failure is undetectable. The first design implication is therefore structural: AI-enabled research workflows need a non-AI instrument capable of measuring what the AI has done.

The term frequency distortions are Messeri and Crockett's illusion of understanding made concrete. The summaries are fluent, they feel complete, but the conceptual center of gravity has shifted. The second design implication is, thus: fluency is not fidelity, and the evaluation criteria applied to AI-enabled research outputs must go beyond coherence and readability to include measurable fidelity to source material.

The checkpoint protocol addresses Ferrario et al.'s conditions for genuine epistemic authority. The researcher who approved a condensation at PEEL's checkpoint exercised authority iteratively, explicitly, and on the record. The instruction encoded in the skill — "Do not infer approval from silence" — is a design principle, not a prompt engineering trick. The third implication is that epistemic authority must be designed into AI-enabled research workflows, not assumed to persist by default.

Taken together, these implications call for reflection on how to evaluate AI-enabled research contributions. The question is not only whether an AI-assisted study is valid, replicable, transparent and scalable. It is whether the researcher remains the epistemic authority throughout: whether their interpretive moves are traceable, their approvals recorded, and their tools' behaviors measurable. PEEL is one answer to that question. It is not the only possible one, and further studies are needed to find out how good it is. But the question itself must become part of how our community designs, conducts, and reviews AI-enabled research in the age of LLMs.

## 7 Conclusion

This commentary invites the IS community to consider how non-AI technologies, combined with AI, can address critical challenges of interpretive, text-based research. We have shown how signs externalized by Claude and Voyant seed abductive reasoning, helping researchers generate creative hypotheses about text sources even before reading them. The process accelerates interpretive research while maintaining the trace of connections and provenance necessary — though not sufficient — to warrant epistemic authority. The researcher drives the process, all the time.

Three design implications follow: deterministic instruments must accompany AI tools; fluency is not fidelity; and epistemic authority must be designed, not assumed. Much remains ahead — scalability, software architecture to lower the costs of tasks deterministic computing can resolve, structured artifacts for research transparency, and the long-term consequences of PEEL in scholarly work. This is only the beginning.

## Acknowledgments

Thanks to everyone who helped and supported us along the way.